\documentclass[10pt,twocolumn,letterpaper]{article}

\usepackage{cvpr}              
\definecolor{cvprblue}{rgb}{0.21,0.49,0.74}
\usepackage[pagebackref,breaklinks,colorlinks,allcolors=cvprblue]{hyperref}
\usepackage{bm}
\usepackage{soul}
\usepackage[accsupp]{axessibility} 

\def\paperID{} 
\def\confName{CVPR}
\def\confYear{2026}

\title{High-Fidelity Mobile Avatars with Pruned Local Blendshapes}

\author{Youyi Zhan$^{1,2}$\footnotemark[2] \quad He Wang$^{2}$ \quad Tianjia Shao$^{1}$\footnotemark[1] \quad Kun Zhou$^{1}$\\
$^1$State Key Lab of CAD\&CG, Zhejiang University \quad $^2$University College London \\
}
\begin{document}

\twocolumn[{%
\renewcommand\twocolumn[1][]{#1}%
\maketitle
\thispagestyle{empty}
\vspace{-6mm}
\begin{center}
     \includegraphics[width= 0.99\linewidth]{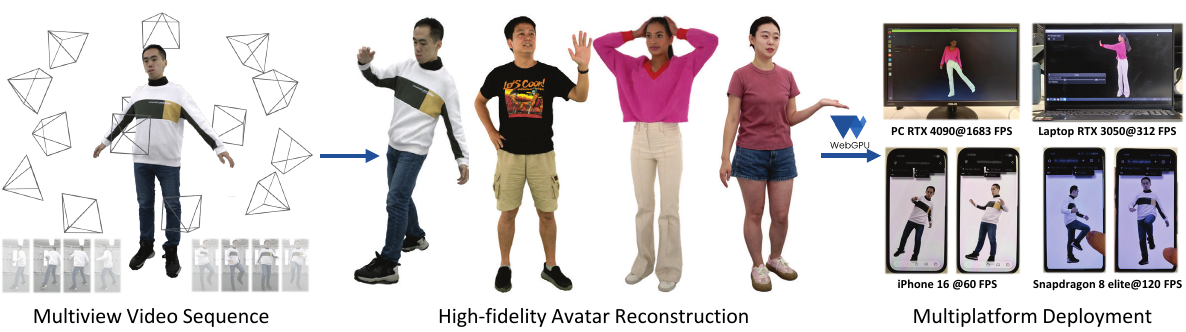}
     \captionof{figure}{Our method creates high-fidelity human avatars from multi-view video. The avatar can run on multiple platforms, including desktop PCs and mobile devices with ultra-fast rendering speed (1683 FPS on an RTX 4090 GPU and 120 FPS on mobile devices).
        }
    \label{fig:teaser}
\end{center}
}]

\maketitle

\renewcommand{\thefootnote}{\fnsymbol{footnote}}
\footnotetext[1]{Corresponding author (tjshao@zju.edu.cn)}
\footnotetext[2]{Work done as a visiting student at University College London}

\begin{abstract}

We propose a method to reconstruct high-fidelity human avatars from multi-view video that can run on mobile devices. Many works can model high-quality Gaussian-based full-body avatars from multi-view video. However, these methods require heavy computation to obtain pose-dependent appearance, making deployment on mobile devices very difficult. Recent methods distill from pretrained models and model pose-dependent nonlinear Gaussian attributes by linearly combining global pose features with blendshapes. Although they can run on mobile devices, they suffer some loss of detail. We observe that nearby Gaussians are often highly correlated within a local region of the body, and can be linearly modeled with less error. Therefore, we use local linear blendshapes in small body parts to capture global nonlinear changes of Gaussian attributes. To further reduce computation and model size, we propose to remove blendshapes for Gaussians whose attributes change little, yielding a minimal blendshape representation. Our method is an end-to-end training method without a pretrained model. To make it run on multiple devices, we implement our method using WebGPU. Experiments show that our method can render high-quality human avatars with better details, and can reach 120 FPS at 2K resolution on mobile devices. Our project page is at \url{https://gapszju.github.io/webavatar/}.

\end{abstract}    

\vspace{-3mm}
\section{Introduction}

The need for high-fidelity 3D human avatar is universal in computer graphics, initially on powerful hardware such as desktop computers and rapidly moving onto portable devices including augmented/virtual reality, mobile devices. Recently, 3D Gaussian Splatting (3DGS~\cite{kerbl20233d}) appears as a new tool for fast construction of realistic human avatars from real images/videos, but it has been mainly used in desktop computers where the computational resources, \eg memory, data bus bandwidth, are abundant, and have not so far achieved similar successes on lightweight hardware platforms such as mobile devices.

Recent research has just started to address this limitation~\cite{chen2025taoavatar,iandola2025squeezeme,shi2025hrm2avatar}, mainly by reducing the needed memory and computation. Inspired by SMPL~\cite{loper2015smpl} using blendshapes for different poses, TaoAvatar~\cite{chen2025taoavatar} and SqueezeMe~\cite{iandola2025squeezeme} uses 3DGS and model the Gaussian attributes as a linear combination of blendshapes and pose features. The main idea behind this is to construct a small basis to represent the distribution of the Gaussian attributes. However, as we show later, their specific strategies are still lacking in terms of training and inference speed and requirements of computation. TaoAvatar and SqueezeMe require pretraining a high-quality large model and then distilling it to a smaller model. Training the original model is slow, as shown by AnimatableGS~\cite{li2024animatable}, and distillation further increases the complexity of the method. Also, both methods use a global pose feature for the whole body to linearly combine all the blendshapes. However, given the nonlinearity of the Gaussian attributes with respect to the pose feature, their linear blending can lead to large errors when representing the Gaussian attributes. Finally, blendshapes take significant storage, memory and computation, especially on bandwidth–limited mobile devices. Although SqueezeMe~\cite{iandola2025squeezeme} attempts to reduce the computation by sharing the same corrective among close-by Gaussians, it causes severe loss of details.

To make 3DGS fully applicable in human avatars on mobile devices, we propose a new method which does not require pre-training, is not demanding in computational resources, and provides good quality. Our key insight is to exploit the locality of Gaussians. The insight stems from the covariance structure of Gaussian attributes. The attributes of nearby Gaussians are often highly correlated (\eg when an area is occluded, a group of Gaussians may all darken together to simulate shadows). This `clustering' effect is especially pronounced on human body surface, where different parts (\eg arms, legs) correspond to different groups of Gaussians which share strong correlations within the group but not necessarily between groups. From the perspective of Principal Component Analysis (PCA) which treats each Gaussian as one sample (\ie a vector of its attributes), if it is done on all the Gaussian attributes together, it is unlikely for PCA to be able to capture the fine-grained covariance structure as data samples are clustered. Instead, if local PCA can be done, then not only can fine-grained local covariance structures be captured, it also enables us to use a few eigenvectors to capture the total variances. Therefore, we partition the body into several parts and the Gaussian attributes in each part are represented as a combination of local pose features and local blendshapes. Unlike SqueezeMe~\cite{iandola2025squeezeme}, we do not obtain the blendshapes by distilling a pre-trained model. We allow our model to learn the blendshapes directly during optimization to avoid the pretrain-distillation process.

Very recently, mmlphuman~\cite{zhan2025real} also follows a similar philosophy and explore to use basis functions to represent Gaussian attributes. Despite the similarity, their method is not suitable for mobile devices as it requires large storage and heavy computation. The key reason is their basis functions contains redundant information. Comparatively, despite also using blendshapes as basis functions, we propose a new pruning technique to learn the minimal set of blendshapes. However, it is non-trivial to do so as the basis is learned, not known a prior. Based on the observation that the dynamic appearance is mainly influenced by a small number of Gaussians, whereas most Gaussians can remain almost unchanged (\eg the appearance of shoes or head stays nearly constant regardless of poses), we can prune the blendshapes for most of the Gaussians, turning them into constant Gaussians. Since we do not know which blendshape should be pruned before training, we train an overparameterized blendshape model, then prune the blendshapes based on the variances of the Gaussian attributes. After further finetuning, this strategy can prune 90\% of the blendshape parameters, significantly reducing memory usage and computational cost.

Through exhaustive evaluation, we demonstrate that our method incurs small computational costs, achieving high-quality rendering, and outperform the existing methods. We implement our method using WebGPU, allowing it to run directly in a web browser as well as on mobile devices. Experiments show that our method is able to model challenging poses and complex appearances, and can achieve 2K resolution at 120 FPS on mobile devices.

Our contributions can be summarized as follows:
\begin{itemize}
\item We propose a new 3DGS method in human avatars which achieves 2K resolution at 120 FPS on mobile devices. 
\item We propose a new pruning strategy to minimize the size of the model while learning leading to a massive speed-up in inference.
\item We analyze the local covariance structure of 3DGS for human avatars, highlighting and provide insights in how to utilize the locality for reducing model size while retaining good quality. 
\item We open-source our implementation. To the best of our knowledge, this is the first open-source full-body human 3DGS method which is capable of real-time and high-quality, pose-dependent appearance rendering on mobile hardware.
\end{itemize}

\section{Related Work}

\subsection{High-fidelity Avatar Rendering}

High-quality human avatars have been explored for a long time. Earlier methods typically rely on mesh with texture mapping~\cite{tong2012scanning, collet2015high, bogo2015detailed, habermann2019livecap}. However, mesh-based representation usually struggles to model complex appearance and geometry. Differentiable volumetric rendering techniques, such as NeRF~\cite{mildenhall2020nerf, mueller2022instant}, enable researchers to model high-fidelity avatar appearance directly from video. Previous works like Nerfies~\cite{park2021nerfies}, HyperNeRF~\cite{park2021hypernerf}, and Li et al.~\cite{li2022neural} model avatar deformations using MLP-based deformation fields. Subsequent methods use tri-plane~\cite{wu2024tetrirf, xu20244k4d} or voxel~\cite{lin2023high, wang2024videorf} or transformer~\cite{shi2024generating} to capture high-frequency space-varying information and render high-fidelity human appearance. 3D Gaussian Splatting (3DGS~\cite{kerbl20233d,wu2024recent}) provides a new representation for avatar modeling. Several methods \cite{zheng2024gps, kwon2025generalizable, liu2025creating} utilize Gaussian rendering to produce high-quality appearance. \cite{wang2025freetimegs, xu2024representing, jiang2024hifi4g, dai20254d, jiang2025topology} can render dynamic scene and represent the avatars in volumetric video format. Although these methods can reconstruct high-quality avatars, they primarily focus on reconstruction and the avatars are not animatable.

\subsection{Animatable Avatar Rendering}

More works focus on reconstructing avatars that can be controlled and animated with given poses. To get realistic rendering, it is essential to model pose-dependent appearance for avatars. Some works~\cite{bagautdinov2021driving, habermann2021real} attempt to model animatable avatars by attaching predicted textures to a deformable mesh to model pose-dependent appearance. NeRF~\cite{mildenhall2020nerf} is also popular to model animatable avatars in recent years \cite{ma2021pixel,liu2024texvocab,peng2021neural,liu2021neural,zheng2022structured,li2022tava,wang2022arah,kwon2024deliffas,weng2022humannerf,jiang2022neuman,yu2023monohuman,peng2021animatable,jiang2023instantavatar,guo2023vid2avatar,geng2023learning,yin2025cyclegaussianavatar,meng2024avatarwild}. These methods typically map points from the observation space to the canonical space and predict color and opacity with input pose as the condition. \cite{li2023posevocab, zheng2023avatarrex, isik2023humanrf} further propose pose encoding methods, enabling high-fidelity and dynamic human appearance. Although high-quality, volumetric representation is slow to render. Even some methods~\cite{deng2024ram, chen2023uv} use neural textures to achieve real-time rendering, they still require substantial compute resources. 3D Gaussian Splatting (3DGS~\cite{kerbl20233d}) achieves higher-quality and faster rendering by rasterizing Gaussians in space. Many methods achieve realistic and real-time results using 3DGS~\cite{saito2024relightable,wen2024gomavatar,moon2024expressive,lei2024gart,kocabas2024hugs,hu2024gaussianavatar,hu2024gauhuman,hu2024expressive,zhan2025interactive, xu2025sequential,lee2025mpmavatar,guo2025vid2avatar}. They typically predict Gaussian attributes with the pose as input to model animatable avatars. Some methods\cite{li2024animatable, shao2024degas, chen2024meshavatar, junkawitsch2025eva, wang2025relightable} further use convolutional networks to decode Gaussian attributes to model dynamic appearance and therefore achieve higher rendering quality compared to other methods. However, these methods still require heavy computation to decode Gaussian attributes. Even though they can run in real time on a desktop GPU, it is still difficult to transfer their models to mobile devices with limited memory bandwidth and computation capacity.

\subsection{Efficient Avatar Rendering}

Improving avatar rendering speed is critical for applications, especially on mobile devices. MoRF~\cite{bashirov2024morf} uses deferred neural rendering to run on mobile at low resolution and low frame rate. SplattingAvatar~\cite{shao2024splattingavatar} attaches Gaussians to mesh template and implements the method on mobile devices, which can render at 30 FPS. However, it does not model dynamic appearance, resulting in suboptimal results. HRM\textsuperscript{2}Avatar~\cite{shi2025hrm2avatar} focuses on monocular capture and mobile running, and therefore cannot model rich dynamic appearance either. Some works recognize the trade-off between decoding dynamic Gaussian attributes and computational cost and seek a balance between the two. A common design is to decode Gaussian attributes using simple blendshape computations. Some recent works~\cite{ma20243d, zhan2025real, zielonka2025gaussian, iandola2025squeezeme, chen2025taoavatar, li2025rgbavatar, aneja2025scaffoldavatar} follow this idea. In particular, ScaffoldAvatar~\cite{aneja2025scaffoldavatar} uses patch-based local expression features for face avatars, but focuses only on the head region. SqueezeMe~\cite{iandola2025squeezeme} and TaoAvatar~\cite{chen2025taoavatar} distill smaller models from convolutional network outputs. Their compact model is expressed by linear blendshapes and can run high-quality avatars in real time on VR headsets. However, they rely on pre-trained models and still fail to capture some details because they use global pose features to linearly express nonlinear Gaussian attributes. Mmlphuman~\cite{zhan2025real} predicts features for each Gaussian to model nonlinear changes, but its heavy blendshapes take much storage and limit the rendering speed. The concept of local blendshapes has been explored in prior facial animation and deformation works~\cite{joshi2006learning, na2011local, tena2011interactive, neumann2013sparse, brunton2014multilinear, wu2016anatomically}. We propose a method with local linear features and a compact blendshape representation that can run high-fidelity animatable full-body avatars on mobile devices.

\section{Method}

Our method reconstructs avatars from multi-view videos, which then can be rendered with high-fidelity appearances under different poses. Similar to previous works~\cite{chen2025taoavatar,li2024animatable,zhan2025real,iandola2025squeezeme}, we first extract the masks of the body and track the SMPL-X~\cite{loper2015smpl} skeleton in each frame. We also employ a canonical template mesh for an avatar. The reconstruction contains two stages. In the first stage (\cref{sec:stage1}), we partition the body into multiple parts, and linearly combine the local pose features and the blendshapes to obtain the Gaussian attributes. In the second stage (\cref{sec:stage2}), we prune the blendshapes to remove as much as redundancy, reducing computational and memory costs. In \cref{sec:train}-\cref{sec:mobile}, we provide implementation details for training and for running the model on mobile devices.

\begin{figure*}[t]
  \begin{center}
    \includegraphics[width=\textwidth]{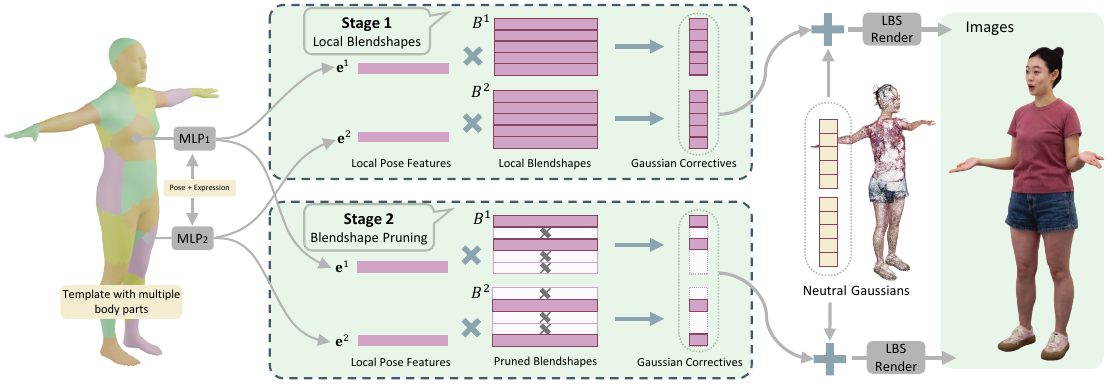}
  \end{center}
  \vspace{-3mm}
  \caption{Pipeline overview. We first partition the template body to multiple parts, and predict a local feature for each part. In the first stage, the local feature is combined with local blendshapes to produce the Gaussian correctives, which are then added to neutral Gaussians to model non-rigid appearance changes. In the second stage, most of the blendshapes are pruned, and we apply correctives to only a small number of Gaussians.
  }
  \label{fig:pipeline}
  \vspace{-3mm}
\end{figure*}

\subsection{Local Linear Blendshapes}
\label{sec:stage1}

Our avatar is represented as $N_g$ Gaussians uniformly distributed on the template mesh. Each Gaussian includes a set of neutral attributes, including rotation $\mathbf{r}$, scale $\mathbf{s}$, color $\mathbf{c}$, opacity $o$, and position $\mathbf{x}$. To get pose-dependent appearance and model non-rigid deformation, we predict Gaussian correctives using local linear blendshapes for a given pose $\bm{\theta}_p$ and expression $\bm{\theta}_e$. Specifically, the body template is partitioned to $N_G$ parts, and Gaussians are grouped to body parts depending on the positions. We follow the position blendshapes of SMPL~\cite{loper2015smpl} and define a local blendshape matrix $B^i$ for the $i$th body part. $B^i$ has shape $[N_{G^i}, N_B, 10]$, where $N_{G^i}$ is the number of Gaussians belonging to the $i$th part, $N_B$ is the dimension of the blendshape, and $10$ represents parameters including rotation in quaternion form, scale and RGB color. Given a local pose feature $\mathbf{e}^i \in \mathbb{R}^{N_B}$, the correctives of rotation, scale and color are predicted by summing up the $N_B$ dimension of blendshape $B^i$,
\begin{equation}
    \{\delta \mathbf{r}^k, \delta \mathbf{s}^k, \delta \mathbf{c}^k | {k \in G^i} \} = B^i \cdot \mathbf{e}^i,
\label{eqn:blendattri}
\end{equation}
where $G^i$ is the Gaussians in the $i$th part, and $\cdot$ denotes the weighted summation over the blendshape dimension.

To predict $\mathbf{e}^i$, we dedicate a small MLP to each body part, which tasks as input the pose and expression, and outputs a local pose feature,
\begin{equation}
    \mathbf{e}^i = \mathsf{MLP}^i(\bm{\theta}_p \oplus \bm{\theta}_e),
\end{equation}
where $\oplus$ is the concatenating operation. To reduce the expression influence on the body, we set the pose to zero vector when the body part is not at head.

Inspired by mmlphuman~\cite{zhan2025real}, we use control points to predict per Gaussian position correctives. We assume there is a global corrective function over the body and we only need to evaluate this function at a set of discrete nodes. More importantly, the evaluation can be locally computed. Specifically,  to determine the nodes, we uniformly sample $N_{s}$ Gaussians from all the Gaussians and treat them as nodes, so each body part contains some nodes. Next, to enable local evaluation, for the $i$th body part, we define a local blendshape matrix $B^i_{s}$ of shape $[N_{G_s^i}, N_B, 3]$ for the nodes belonging to this part, where $N_{G_s^i}$ is the number of nodes in the $i$th part. As a result, each node's position corrective $\delta \mathbf{x}_s$ can be obtained by summing up the blendshapes, similar to \cref{eqn:blendattri}:
\begin{equation}
\{\delta \mathbf{x}_s^k | k \in G^i_s\} =  B^i_s \cdot \mathbf{e}^i .
\end{equation}
Finally each Gaussian's position corrective $\delta \mathbf{x}$ is interpolated from the three nearest nodes,
\begin{equation}
\delta \mathbf{x} = \frac{\sum_j \alpha_j \delta \mathbf{x}^j_s}{\sum_j \alpha_j},
\end{equation}
where $\alpha_j = 1 / \| \mathbf{x} - \mathbf{x}^j_s \|$ is the interpolation coefficent and $j$ is the index of the three nodes nearest to $\mathbf{x}$. 

Similar to TaoAvatar~\cite{chen2025taoavatar}, we don't use high-order SH coefficients to model view-dependent color as the body usually tends to be diffuse. We also do not impose correctives on opacity and let it to be a constant attribute. After applying the correctives, the Gaussians are deformed to posed space and splatted as 2D primitives to render an image with Gaussian rasterizer~\cite{kerbl20233d,ye2025gsplat}.

\subsection{Blendshape Pruning}
\label{sec:stage2}

Although linearly combining the blendshapes to obtain the correctives is effective as it doesn't involve complicated neural network computation, the model still takes extra storage and memory, because the blendshapes are usually several times as big as the Gaussian attributes. Therefore, we seek a minimal blendshape representation by removing redundant basis functions. We observe that some parts of an avatar change little across different views and poses, such as the shoes. This means the Gaussians at these places do not need pose-dependent correctives and the corresponding blendshapes can be pruned.

\begin{figure}[tbp]
  \begin{center}
    \includegraphics[width=0.95\columnwidth]{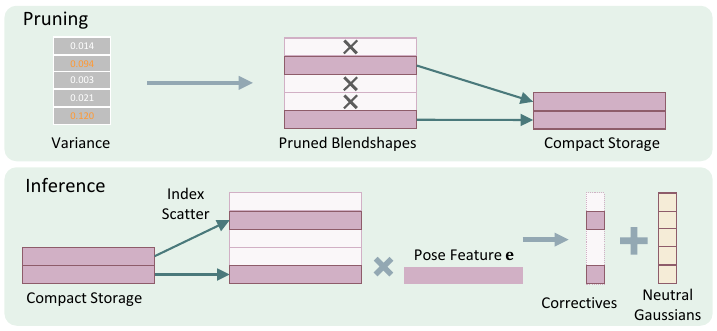}
  \end{center}
  \vspace{-3mm}
  \caption{Illustration of blendshape pruning. 
  }
  \label{fig:prune}
  \vspace{-5mm} 
\end{figure}

However, because we do not know which blendshapes should be kept as we do not know what they are before training, we first train an overparameterized blendshape model and then prune it. Specifically, after the first-stage training, we collect the output correctives of all the training poses and expressions, and compute the variances of rotation, scale and color correctives for each Gaussian, 
\begin{equation}
\begin{aligned}
\hat{r} &= Var(\{\delta \mathbf{r}_p\}) \\
\hat{s} &= Var(\{\delta \mathbf{s}_p\}) \\
\hat{c} &= Var(\{\delta \mathbf{c}_p\}), 
\end{aligned}
\label{eqn:variance}
\end{equation}
where $p \in \Theta$ is the training pose and expression, and $\Theta$ contains all the training poses and expressions. $\{\delta \mathbf{r}_p\}, \{\delta \mathbf{s}_p\}, \{\delta \mathbf{c}_p\}$ are the collected attributes of all the poses for a Gaussian. We then keep the blendshape for Gaussians whose attributes exhibit large variance, and prune the blendshape component for Gaussians with small variance. For each attribute type, we independently keep the blendshapes of the top $N_P$ Gaussians with the largest variance and prune the others, where $N_P$ is applied separately to rotation, scale and color, and $N_P$ is a number much smaller than the number of Gaussians $N_g$. This means the rotation, color, and scale attributes will each retain $N_P$ blendshapes after pruning. This enables us to store the sparse blendshape in a compact form to reduce storage and memory usage. \cref{fig:prune} provides an illustration of the pruning process. We note that the rotation, scale and color blendshapes are pruned separately based on their variances. After this pruning, we further fine-tune the model.

Since the pruning process deletes the correctives and can affect the appearance, we add a constraint on the correctives in the first stage for mitigation. This constraint is to encourage the correctives to be as small as possible, so that the pruning process will not greatly affect the rendering results,
\begin{equation}
\mathcal{L}_{cst} = \lambda_{\delta\mathbf{r}} \| \delta \mathbf{r} \|_1 + \| \delta \mathbf{s} \|_1 + \lambda_{\delta\mathbf{c}} \| \delta \mathbf{c} \|_1
\end{equation}
where $\lambda_{\delta\mathbf{r}} = 0.02$, $\lambda_{\delta\mathbf{c}} = 0.002$. During the fine-tuning stage, the loss is not used.

\subsection{Implementation Details}
\label{sec:train}

At the beginning, we initialize $N_g=200K$ Gaussians on the template mesh. We use SMPL-X~\cite{pavlakos2019expressive} mesh as the template for most of the cases, and construct the template following~\cite{li2024animatable} for loose clothing. To partition the body, we sample $N_G=256$ points on the template mesh using Poisson-disk sampling, and assign each of the $N_g$ Gaussians to its nearest sampled point. All Gaussians mapped to the same point form one partition. The template mesh is only used for initialization, and once the Gaussians are sampled and grouped to different body parts, we no longer use the mesh. We set $N_G=256, N_B=16, N_s=10K$ and $N_P=20K$ in our experiments. The Gaussian position $\mathbf{x}$ is fixed after being sampled on the template mesh, and we predict $\delta \mathbf{x}$ to compensate the non-rigid displacement under different poses. $\bm{\theta}_e$ consists of the first 10 FLAME~\cite{li2017learning} expression components and $\bm{\theta}_p$ is 63-dimensional, containing the axis-angle rotations of the main joints.

During training, we employ the same loss function as AnimatableGS~\cite{li2024animatable}, which includes L1 loss, LPIPS loss. We also apply a scale loss to prevent the Gaussian from growing too large. The final loss is,
\begin{equation}
\mathcal{L} = \mathcal{L}_{1} + \lambda_{lpips} \mathcal{L}_{lpips} + \mathcal{L}_{scale} + \mathcal{L}_{cst},
\end{equation}
where $\lambda_{lpips}$ is set to $0.1$. The constraint loss $\mathcal{L}_{cst}$ is only applied in the first stage, and will not be used after pruning. We train the first stage for 200K iterations, and second fine-tuning stage for 80K iterations. Training batch size is set to 4. The whole training takes 7.5 hours on an RTX 4090 GPU.

\subsection{Running on Mobile Devices}
\label{sec:mobile}
To deploy our model to mobile devices, we quantize all floating-point parameters, including MLP, blendshapes and attributes, to \textsf{float16}, and export some integer parameters to \textsf{int16}. After compression, the model size is just about 19.4 MB and can be easily downloaded on various types of devices. We implement our visualization program using Rust and WebGPU, which can be compiled to a native executable on desktop system (e.g., Windows, macOS and Linux), or a webpage that can be opened on the supported browser. WebGPU is a modern graphics interface that can fully utilize the GPU for rendering and parallel computing. All tests on the mobile devices are conducted by opening the webpage on the browser. We use WebGPU’s compute shaders for MLP inference, combining the blendshapes and animating the avatar with linear blend skinning. Then the Gaussians are rendered using a rasterizer from c3dgs~\cite{niedermayr2024compressed}. This rasterizer performs efficient Gaussian sorting and blending using WebGPU. The whole process is fully accelerated by GPU hardware, and can reach 1683 FPS on an RTX 4090 GPU, or 120 FPS on mobile devices with 2K resolution.

\begin{figure*}[t]
  \begin{center}
    \includegraphics[width=0.95\textwidth]{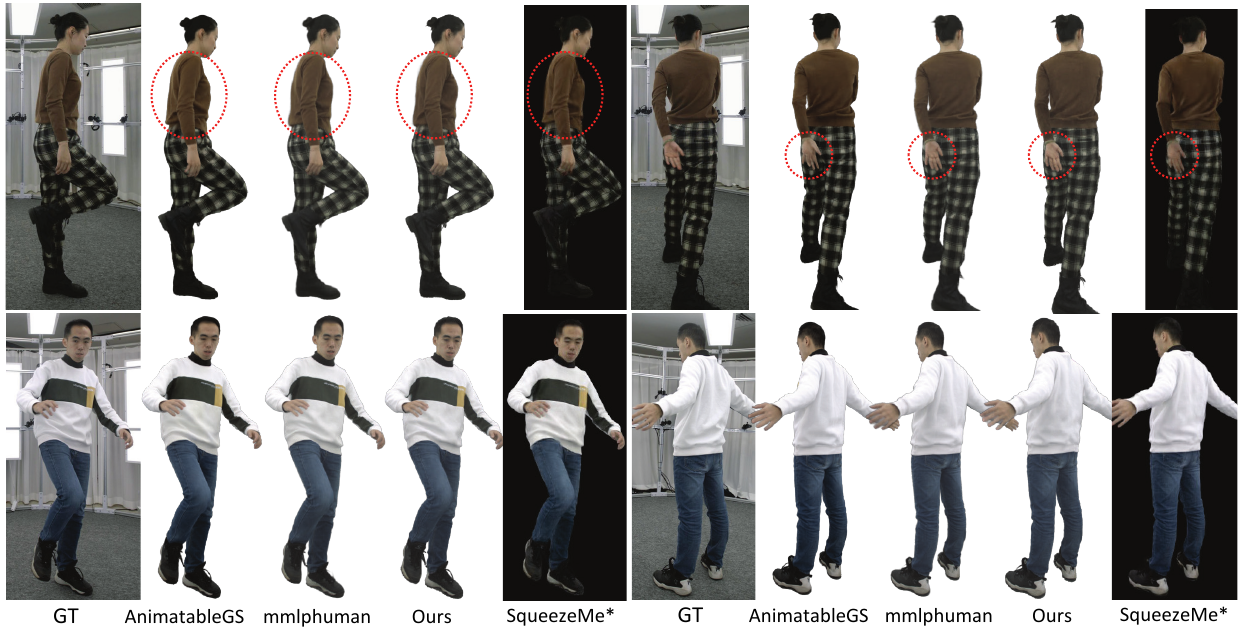}
  \end{center}
  \vspace{-3mm}
  \caption{Comparison on AvatarRex dataset. All images are rendered with novel view and novel pose. *The results of SqueezeMe are taken from the appendix of their paper.
  }
  \label{fig:compare}
  \vspace{-3mm}
\end{figure*}

\section{Experiments}

\textbf{Dataset.} We employ four datasets. \textbf{AvatarRex}~\cite{zheng2023avatarrex}: Each sequence contains videos captured using 16 cameras with 2K resolution. The avatar performs various poses and exhibits  rich pose-dependent appearance. \textbf{TalkingBody4D}~\cite{chen2025taoavatar}: This dataset contains videos of the avatar standing in front of the cameras and speaking, accompanied by facial expressions and hand gestures. Each sequence contains 60 cameras with 2K resolution. \textbf{ActorsHQ}~\cite{isik2023humanrf}: Each sequence contains about 1,000-2,000 frames and 160 cameras. We use only about 40 cameras. We also follow previous works and use 4x down-sampled images. \textbf{DREAMS-Avatar}~\cite{shao2024degas}: It contains videos in which the avatar performs different poses with rich expressions. Each sequence has 32 viewpoints at 2K resolution and about 1,000–2,000 frames. We used the datasets’ provided SMPL-X registrations for the above datasets. Since ActorsHQ does not contains the skeleton registration, we use the registration provided by AnimatableGS~\cite{li2024animatable}.

\begin{figure}[t]
  \begin{center}
    \includegraphics[width=\columnwidth]{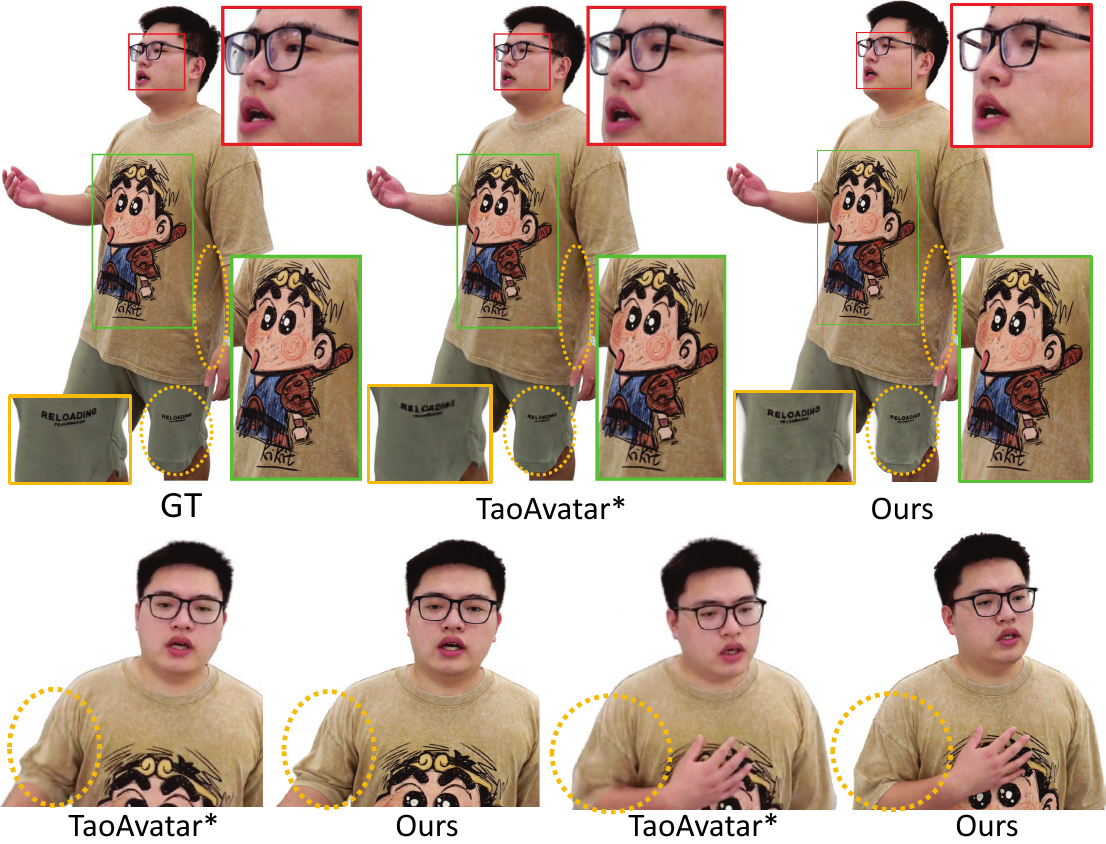}
  \end{center}
  \vspace{-3mm}
  \caption{Qualitative comparison with TaoAvatar~\cite{chen2025taoavatar}. *The results of TaoAvatar are taken from their paper and video.
  }
  \label{fig:comparetao}
  \vspace{-3mm} 
\end{figure}

\noindent \textbf{Platform and Performance Evaluation.} We test the performance on three platforms: a laptop with an RTX 3050 GPU, a PC with an RTX 4090 GPU and an Android phone equipped with a Snapdragon 8 Elite chip. Our model can be run using the PyTorch implementation, or the executable compiled from Rust code. Since our method can run very fast and the python overhead is heavy, to fully utilize platform performance, we use the compiled executable for all performance experiments unless otherwise specified.

\subsection{Comparison}

We compare with AnimatableGS~\cite{li2024animatable}, mmlphuman~\cite{zhan2025real}, TaoAvatar~\cite{chen2025taoavatar} and SqueezeMe~\cite{iandola2025squeezeme} for high-quality multi-view avatar reconstruction. We emphasize that both TaoAvatar and SqueezeMe do not release the code or model for training or evaluation. Therefore, we follow their setting and take quantitative results and images from their papers for comparison. All results taken from their paper are marked with an asterisk (*).

Since SqueezeMe reports some results on the AvatarRex dataset, we follow SqueezeMe’s experimental setup as closely as possible. Specifically, we test on \textsf{lbn2} and \textsf{zzr} identities, training with 14 cameras and reserving two cameras for novel-view evaluation. The final 500 frames of each sequence are held out for novel pose evaluation test data. We overlay the rendered avatar on the groundtruth image, and compute L1, PSNR, SSIM, and LPIPS as evaluation metrics. The metrics are only calculated inside the bounding box of the groundtruth masks. We evaluate using novel view and novel poses. We compare with AnimatableGS, mmlphuman and SqueezeMe using this setting, and the results are presented in \cref{table:compare}. AnimatableGS achieves the best novel-pose animation quality but is computationally heavy and slow. Mmlphuman produces results comparable to ours, but its large model size makes mobile deployment challenging. Under the same experimental settings, our results outperform SqueezeMe. We also provide qualitative results in \cref{fig:compare}. Our results are comparable to SqueezeMe, but with more details in some cases (red circles), and are also comparable to mmlphuman and AnimatableGS.

\newcommand{\iconyes}{\raisebox{-0.2ex}{\includegraphics[height=1em]{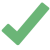}}}
\newcommand{\iconno}{\raisebox{-0.2ex}{\includegraphics[height=1em]{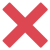}}}

\begin{table}[t]\
\centering
\resizebox{\columnwidth}{!}{
\begin{tabular}{l|cccccc}
\toprule
          & L1$\downarrow$      & PSNR$\uparrow$   & SSIM$\uparrow$   & LPIPS$\downarrow$  & FPS\textsubscript{4090}$\uparrow$ & On Mobile  \\ \hline
AnimatableGS~\cite{li2024animatable}     & 0.02270 & 25.508 & 0.8655 & 0.1550 & 16       & \iconno             \\
mmlphuman~\cite{zhan2025real} & 0.02371 & 25.276 & 0.8617 & 0.1573 & 315      & \iconno            \\
Ours      & 0.02343 & 25.346 & 0.8606 & 0.1576 & 1683     &  \iconyes            \\ \hline
SqueezeMe*~\cite{iandola2025squeezeme} & 0.059   & 20.051 & 0.849  & 0.158  & --      & \iconyes              \\
TaoAvatar*~\cite{chen2025taoavatar} & -- & --  & -- & --            & 156      &  \iconyes             \\ \bottomrule
\end{tabular}
}
\vspace{-2mm}
\caption{Quantitative comparison with high-quality multi-view methods. We experiment on AvatarRex dataset and follow SqueezeMe's experiment settings. *The results of SqueezeMe~\cite{iandola2025squeezeme} and TaoAvatar~\cite{chen2025taoavatar} are taken from their paper.}
\label{table:compare}
\vspace{-3mm}

\end{table}

TaoAvatar~\cite{chen2025taoavatar} evaluates on TalkingBody4D. Therefore, we also run experiments on the same dataset. We use four sequences. For each sequence we hold out one camera for testing and train with the remaining cameras. The selected sequence and test viewpoint match those in the figure 3 in TaoAvatar paper. In each sequence we also reserve last 200 frames for novel pose testing. Metrics are computed over the entire image. \cref{table:comparetao} shows the quantitative results. Our method achieves better results for most of the metrics. \cref{fig:comparetao} also shows novel-view comparisons with TaoAvatar. The images are picked from its paper and supplementary video. Our method is able to captures more details, such as small text on the pants and wrinkles on the clothing. In summary, among the methods that can run on mobile devices (TaoAvatar and SqueezeMe), our method can produce better results on the same data case.

We also present the performance of the methods in \cref{table:compare}. The rendering resolution is 2K. Our design can decode Gaussian attributes under different poses in very little time and can run extremely fast on a 4090 GPU, outperforming other methods in efficiency.

\begin{table}[]
\centering
\resizebox{\columnwidth}{!}{
\begin{tabular}{l|ccc|ccc}
\toprule
          & \multicolumn{3}{c|}{Novel View} & \multicolumn{3}{c}{Novel Pose and Expression} \\ \cline{2-7} 
          & PSNR$\uparrow$     & SSIM$\uparrow$     & LPIPS$\downarrow$     & PSNR$\uparrow$         & SSIM$\uparrow$          & LPIPS$\downarrow$          \\ \hline
TaoAvatar*~\cite{chen2025taoavatar} & 33.81    & 0.9689   & 0.06437   & 28.38        & 0.9389        & 0.08874        \\
Ours      & 34.44    & 0.9771   & 0.03642   & 27.90        & 0.9395        & 0.05582  \\  
\bottomrule
\end{tabular}
}
\vspace{-2mm}
\caption{Quantitative comparison with TaoAvatar on TalkingBody4D dataset.}
\label{table:comparetao}
\vspace{-2mm}

\end{table}

\begin{table}[]
\centering
\resizebox{0.85\columnwidth}{!}{
\begin{tabular}{l|cccc}
\toprule
              & L1$\downarrow$       & PSNR$\uparrow$    & SSIM$\uparrow$    & LPIPS$\downarrow$   \\ \hline
Ours          & 0.01488 & 29.405 & 0.9207 & 0.1074 \\
Global feature 16 & 0.01777 & 27.645 & 0.8978 & 0.1276 \\
Global feature 64 & 0.01744 & 27.913 & 0.8988 & 0.1270 \\
No pruning    & 0.01471 & 29.342 & 0.9188 & 0.1076 \\
\bottomrule
\end{tabular}
}
\vspace{-2mm}
\caption{Ablation study on several designs. This table is evaluated on the training pose and novel view on AvatarRex dataset.}
\label{table:ablation}
\vspace{-2mm}
\end{table}

\begin{table}[]
\centering
\resizebox{0.8\columnwidth}{!}{
\begin{tabular}{l|ccc}
\toprule
           & FPS\textsubscript{3050} & GPU Mem(MB) & Model Size(MB) \\ \hline
Ours       & 312      & 105     & 19.5     \\
No Pruning & 191      & 155     & 72.4       \\
\bottomrule
\end{tabular}
}
\vspace{-2mm}
\caption{Comparison with no pruning design. All render resolution is 2K. }
\label{table:ablationnp}
\vspace{-2mm}

\end{table}

\begin{figure}[t]
  \begin{center}
    \includegraphics[width=0.9\columnwidth]{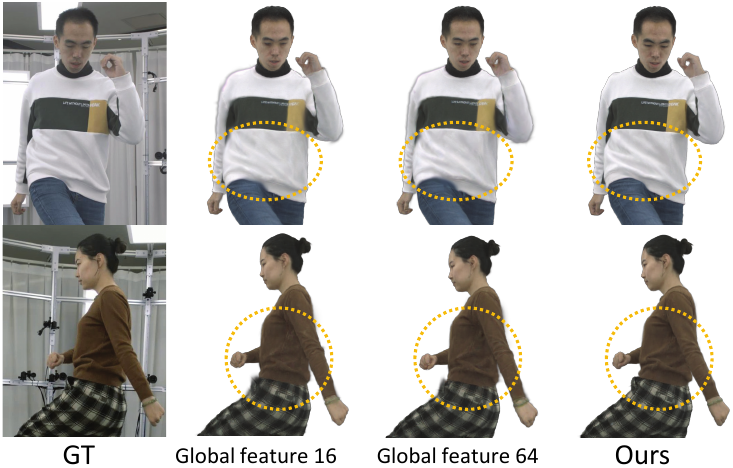}
  \end{center}
  \vspace{-4mm}
  \caption{Ablation study on local and global features.
  }
  \label{fig:ablationlocal}
  \vspace{-3mm} 
\end{figure}

\begin{figure}[t]
  \begin{center}
    \includegraphics[width=0.99\columnwidth]{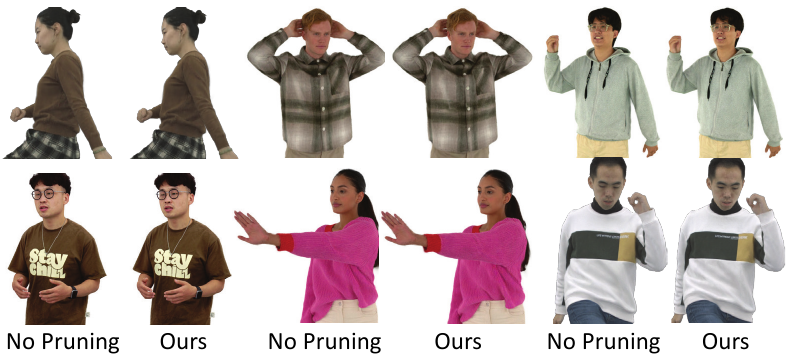}
  \end{center}
  \vspace{-4mm}
  \caption{Qualitative comparison with no pruning design. The results of no pruning are almost identical to ours.
  }
  \label{fig:ablationprune}
  \vspace{-3mm} 
\end{figure}

\begin{figure*}[t]
  \begin{center}
    \includegraphics[width=0.95\textwidth]{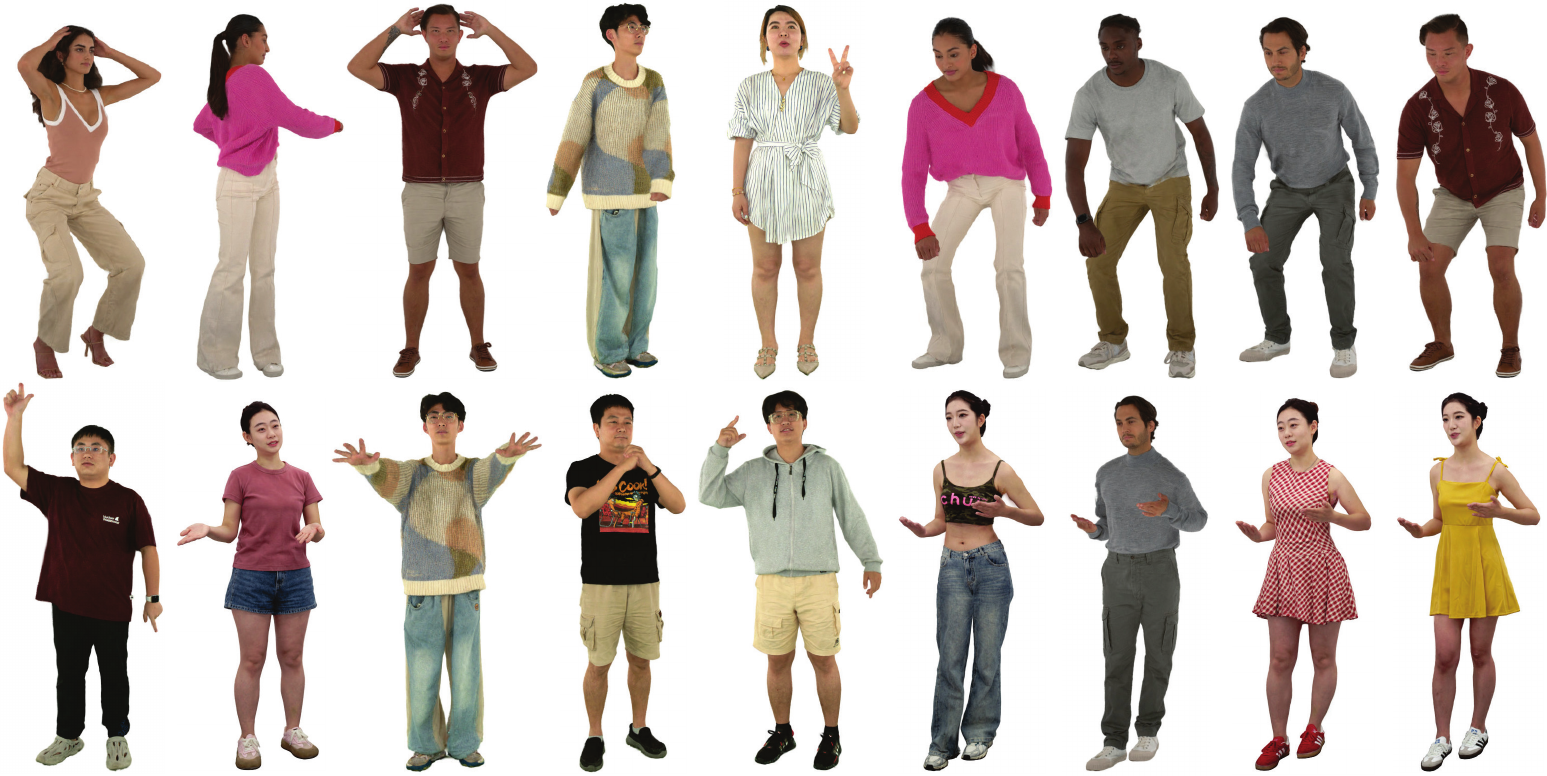}
  \end{center}
  \vspace{-3mm}
  \caption{Our method can reconstruct high-fidelity human avatars with various appearance and poses. The avatars can be animated under novel poses.
  }
  \label{fig:gallery}
  \vspace{-6mm}
\end{figure*}

\begin{figure}[ht]
  \begin{center}
    \includegraphics[width=0.95\columnwidth]{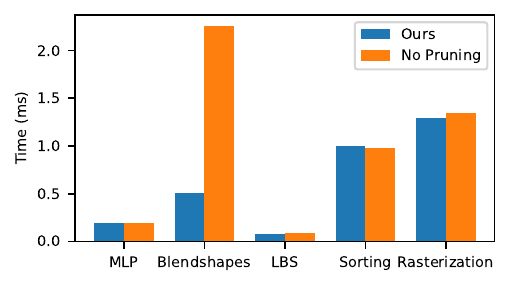}
  \end{center}
  \vspace{-5mm}
  \caption{Runtime of each part per frame. The data is collected on a 3050 GPU with 2K rendering resolution.
  }
  \label{fig:frametime}
  \vspace{-5mm} 
\end{figure}

\subsection{Ablation Study}

\textbf{Local Linear Blendshape.} Both SqueezeMe~\cite{iandola2025squeezeme} and TaoAvatar~\cite{chen2025taoavatar} use a global pose feature to combine blendshapes to obtain Gaussian attributes. We show that modeling Gaussian pose-dependent attributes with local linear blendshapes performs better. We use an MLP that takes the pose and expression as input and output a pose feature vector of length 16, which is used to combine all Gaussian blendshapes. We evaluate the design on novel views of AvatarRex dataset. \cref{table:ablation} shows the results. The global pose feature performs far worse. We also test using a pose feature of length 64, which is the same setting used by SqueezeMe, and the results won't get better. \cref{fig:ablationlocal} also shows that using a global pose feature loses more details. Although SqueezeMe mentions that training the global linear model from scratch performs worse than distilling from a trained model, our method is end-to-end and does not require training a heavy convolutional network like TaoAvatar and SqueezeMe.

\noindent \textbf{Blendshape Pruning.} Pruning blendshapes may reduce the model's learning capacity, since most of the Gaussians' attribute will be constant and won't change along with the pose. But we find its impact on rendering results is limited. We compare with a model without the pruning and fine-tuning stage. We note that the first stage model is converged. The quantitative results are given in \cref{table:ablation}. We show that after pruning and fine-tuning our design can achieve comparable results. This indicates that pruning blendshapes does not severely impair the model's ability to learn appearance. \cref{fig:ablationprune} also presents the differences between the two designs. Our pruning design produces results similar to the ones without pruning.

Besides delivering comparable quality, our design further reduces model size and improves speed. We report storage and runtime metrics in \cref{table:ablationnp}. Because we prune 90\% of the blendshapes, the overall model size and memory footprint are substantially reduced. Although the FPS improvement is not as large as the reduction in model size, we attribute this to a rendering bottleneck. \cref{fig:frametime} shows the average time per frame for the two designs. The sorting and rasterization take most of the time. Compared with no pruning, our design still greatly reduces the time of blendshape combination (0.52 ms vs. 2.26 ms).

\subsection{Runtime Performance and Application}
\cref{fig:frametime} shows that compared to the overall time for one frame, the time of decoding Gaussian attributes (MLP + Blendshapes + LBS) is very short. The bottleneck lies in sorting and rendering the Gaussians. Our method can achieve 1683 FPS on 4090 GPU and 120 FPS on mobile phone. Due to browser limitation, the FPS can only reach to the maximum refresh rates of the screen on the phone. Our method decodes Gaussian attributes every frame and can still saturate display refresh rates, demonstrating very high efficiency. Please refer to the supplemental video for the interactive demo. \cref{fig:gallery} provides the examples of the avatars reconstructed by our method. We also provide the applications in the supplementary material that allow users to run the digital avatar on their own devices.

\section{Conclusion}

We propose a lightweight avatar rendering method that can model high-quality dynamic appearance. Compared to previous methods capable of running on mobile devices, our method can be trained end-to-end and does not require pretraining a large convolutional network. We model globally nonlinear Gaussian attributes using local linear modeling to capture finer detail. Blendshape pruning further removes redundant model parameters, increasing speed and reducing model size. We implement our method with WebGPU and run it in the browser. With just a click, a lifelike digital avatar can be brought to anyone, enabling immersive, interactive, and personalized experiences that enhance communication between the physical and virtual worlds.

\clearpage

\section*{Acknowledgment}

We thank the reviewers for their insightful comments and Doğa Yılmaz for the assistance with this paper. This work is supported in part by NSF China under Grant 62322209 and Grant 62421003, and the XPLORER PRIZE.

{
    \small
    \bibliographystyle{ieeenat_fullname}
    \bibliography{main}
}

\clearpage
\maketitlesupplementary

\section{Attribute PCA Analysis}

To demonstrate that the attributes of local Gaussians are highly correlated, we perform PCA on Gaussian attributes. We use color as the example. We collect the output RGB color attributes from AnimatableGS~\cite{li2024animatable}. There are $F = 1500$ frames and the model has $N_g = 370K$ Gaussians, so all colors form a large matrix. We subtract each Gaussian’s mean color to obtain the color corrective matrix $X$ with size $[F, 3N_g]$.

For SqueezeMe~\cite{iandola2025squeezeme} and TaoAvatar~\cite{chen2025taoavatar}, they use a global pose feature to combine all the blendshapes. That is, they represent the full color matrix $X$ using a blendshape matrix of size $[N_B, 3N_g]$, where $N_B$ is the dimension of the blendshape space used in their paper. We use PCA on all the Gaussians (global PCA) to measure how large the approximation error of this representation is. We reduce the matrix $X$ to $[N_B, 3N_g]$, using the first $N_B$ dimensions, then project it back to original space $[F, 3N_g]$ as a reconstruction. To measure the reconstruction error, we compute the L1 error between $X$ and its reconstruction. We test with $N_B = 16$ and $64$ and report the error in \cref{table:pcal1}. We note that $N_B = 16$ is the parameter used in our paper and $N_B=64$ is the same setting used in SqueezeMe~\cite{iandola2025squeezeme}.

For our design, we group Gaussians by their positions according to the different parts of the body, and use local pose feature to combine the local blendshape within each group. To measure the approximation error of this representation, we group the Gaussians from AnimatableGS~\cite{li2024animatable} in the same way as we do in the paper. Then we form many color matrices $\{X_k\}$, one for each group, perform PCA on each group (local PCA) to reduce the dimension, then project them back to the original space as reconstructions. We also compute the L1 error between the original color matrices and their reconstructions. The results are shown in \cref{table:pcal1}. The local PCA produces lower error, indicating that grouping Gaussians by positions yields better reconstruction error.

\begin{figure}[t]
  \begin{center}
    \includegraphics[width=\columnwidth]{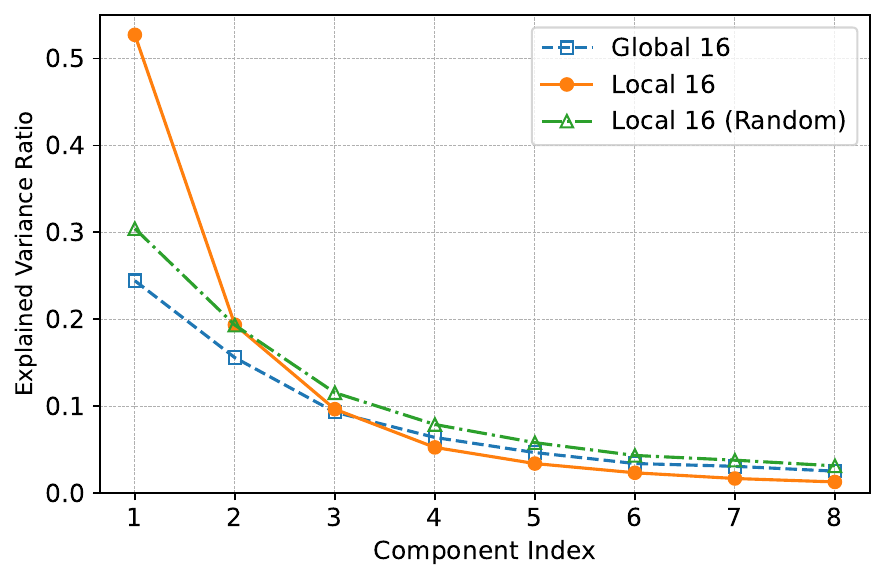}
  \end{center}
  \vspace{-6mm}
  \caption{Explained variance ratio for local and global PCA.}
  \label{fig:variance}
  \vspace{-1mm} 
\end{figure}

We also report the explained variance ratio to show how concentrated the principal components are for local and global PCA. In PCA, the explained variance ratio is the fraction of the dataset’s total variance that a principal component accounts for. A higher explained variance ratio means the component carries more of the original data’s information. The ratio for local PCA is the average explained variance ratios computed for all matrices $\{X_k\}$. We present the results in \cref{fig:variance}. The ratios of local PCA are more concentrated in the first few components, further demonstrating that local Gaussians’ attributes are more correlated and can be captured with fewer blendshapes to capture the total variance.

We further emphasize that this correlating property can only be captured by grouping based on locality. To demonstrate this, we group Gaussians to random groups, perform local PCA and compute errors, as shown in \cref{table:pcal1}. We also report the explained variance ratios in \cref{fig:variance}. This random grouping design produces large errors and poorly concentrated information in the principal components.

The above experiments shows that using local blendshapes to model Gaussians can produce lower error. Unlike SqueezeMe~\cite{iandola2025squeezeme}, which obtains blendshapes by distilling the output of a convolutional network, we allow our model to learn the local blendshapes during optimization.

\begin{table}[t]
\centering
\resizebox{0.99\columnwidth}{!}{
\begin{tabular}{l|cccc}
\toprule
         & Global 16 & Global 64 & Local 16 & Local 16(Random) \\ \hline
L1 Error & 0.0159    & 0.0092    & 0.0042   & 0.0158         \\
\bottomrule
\end{tabular}
}
\caption{Error analysis for global and local PCA.}
\label{table:pcal1}
\end{table}

\section{Pruning Visualization}

\begin{figure*}[t]
  \begin{center}
    \includegraphics[width=0.9\textwidth]{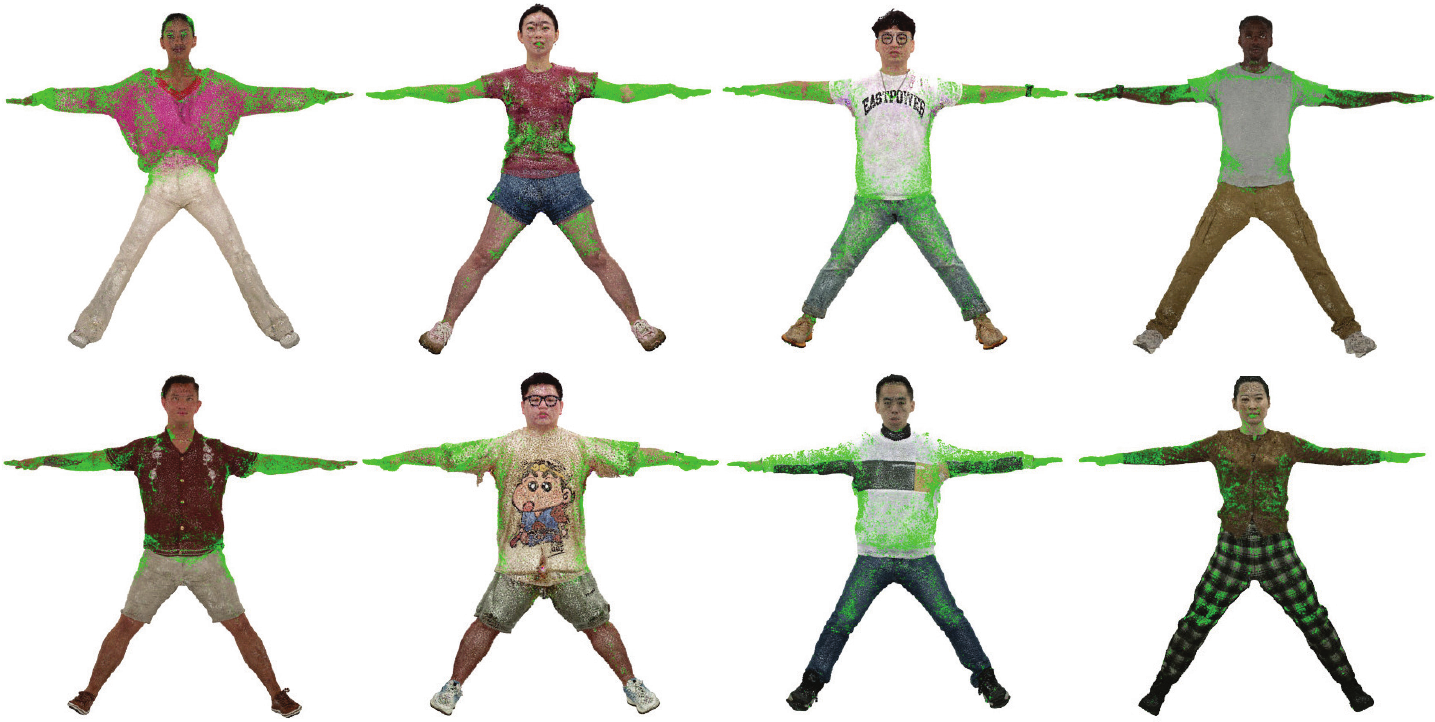}
  \end{center}
  \vspace{-3mm}
  \caption{Visualization of Gaussians that contain the color blendshapes. The green points are the Gaussians whose color blendshapes have not been pruned.
  }
  \label{fig:visblendshape}
  \vspace{-3mm}
\end{figure*}

We visualize the Gaussians on the body that keep color blendshapes across different identities, as shown in \cref{fig:visblendshape}. These Gaussians are mainly located in regions with rich pose-dependent appearance, such as the clothing wrinkles and arms. Other body parts, like lower legs and shoes, show little appearance change along with pose change. Therefore, the blendshapes of Gaussians at these places can be pruned. We note that as the avatar’s arm poses change, the lighting and appearance of the arms also change. Therefore many Gaussians' blendshapes are retained on the arms to model this variation. 

\section{Implementation Details}

\noindent \textbf{MLP Architecture.} The MLP on each part of the body contains three hidden layer. Each layer's size is 64. For 256 MLPs, all the parameters take about 7MB storage with \textsf{float16} quantization. 

\noindent \textbf{Novel Pose Animation.} We follow AnimatableGS~\cite{li2024animatable} and mmlphuman~\cite{zhan2025real} to use PCA to project the novel pose and novel expression into the distribution of the training poses and expressions to prevent unpleasant rendering results.

\section{Partition Number Ablation}

\begin{figure}[t]
  \begin{center}
    \includegraphics[width=\columnwidth]{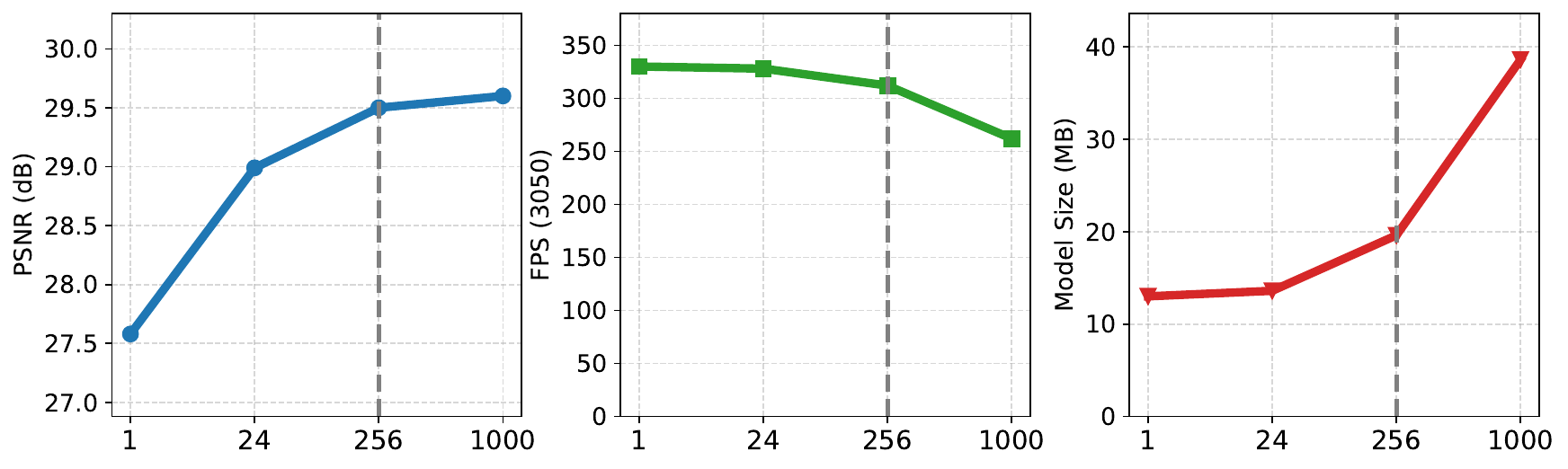}
  \end{center}
  \vspace{-6mm}
  \caption{Ablation on the number of body partitions $N_G$.}
  \label{fig:partitionnum}
  \vspace{-1mm}
\end{figure}

We ablate the number of body partitions $N_G$ to evaluate its effect on quality, speed, and model size. \cref{fig:partitionnum} shows how PSNR, FPS, and model size vary with the partition count. The data is measured on the training pose and novel view of \textsf{avatarrex\_zzr} sequence. Rather than dividing the body into 24 parts based on the SMPL joints, we partition it into more parts to capture finer-grained local structure and improve reconstruction quality. However, increasing the number of partitions also increases model size and reduces rendering speed. A balanced choice among model size, rendering speed, and visual quality is  $N_G=256$ partitions.

\section{Pruning Threshold $N_P$ Ablation}

\begin{figure}[t]
  \begin{center}
    \includegraphics[width=\columnwidth]{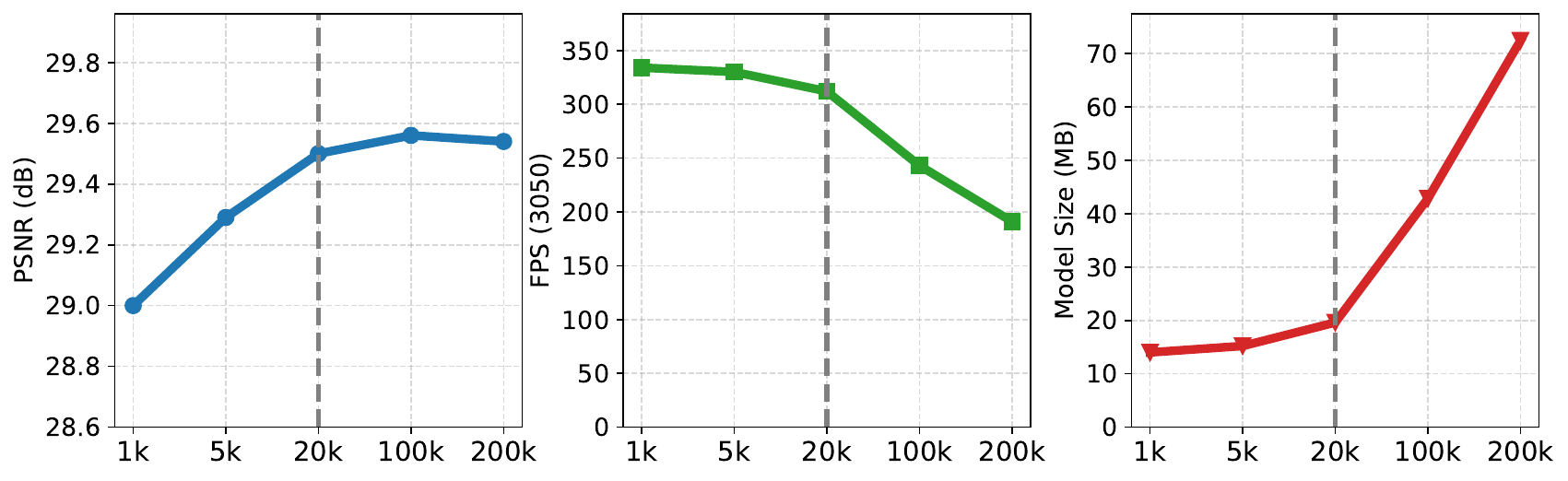}
  \end{center}
  \vspace{-6mm}
  \caption{Ablation on the pruning threshold $N_P$.}
  \label{fig:npnum}
  \vspace{-1mm}
\end{figure}

We ablate the pruning threshold $N_P$ that controls how many Gaussians retain their blendshapes after pruning. \cref{fig:npnum} shows how PSNR, FPS, and model size vary with $N_P$. To balance quality, model size, and rendering speed, we empirically choose $N_P = 20K$ in our experiments. We note that $N_P$ is applied separately to each attribute type, i.e., the rotation, color, and scale attributes will each retain 20K blendshapes after pruning.

\section{Mobile Performance Analysis}

\begin{figure}[t]
  \begin{center}
    \includegraphics[width=\columnwidth]{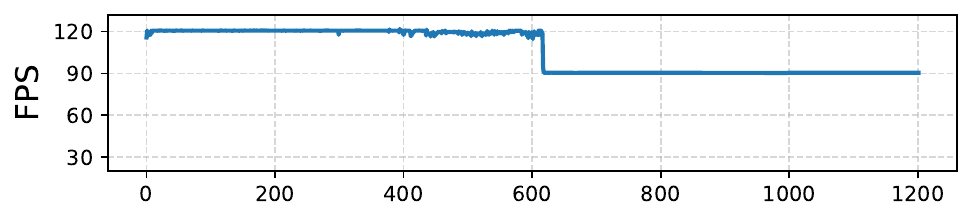}
  \end{center}
  \vspace{-6mm}
  \caption{FPS over 20 minutes of continuous rendering on a mobile device (Snapdragon 8 Elite).}
  \label{fig:fpstime}
  \vspace{-1mm}
\end{figure}

To evaluate the stability of rendering performance on mobile devices, we measure the FPS over a continuous 20-minute session, as shown in \cref{fig:fpstime}. On the Snapdragon 8 Elite device, the rendering maintains 120 FPS for the first 10 minutes before thermal throttling reduces it to approximately 90 FPS for the remainder of the session. We also test on mid-range phones with Snapdragon 8 Gen 1 and Google Tensor G3 chips (released 4 and 3 years ago respectively), both achieving 60 FPS, demonstrating that our method is not limited to high-end devices.

\section{SplattingAvatar Speed Comparison}

SplattingAvatar~\cite{shao2024splattingavatar} cannot model pose-dependent appearance so we didn't include  in the quality evaluation. For speed, SplattingAvatar achieves 50 FPS\textsubscript{3050}, while our method achieves 312 FPS\textsubscript{3050}. When not predicting dynamic Gaussians, our method reaches 410 FPS\textsubscript{3050} (dynamic-Gaussian-prediction time / total time = 0.695 ms / 3.053 ms per frame), indicating that predicting dynamic appearance does not add significant overheads.

\begin{figure}[t]
  \begin{center}
    \includegraphics[width=0.6\columnwidth]{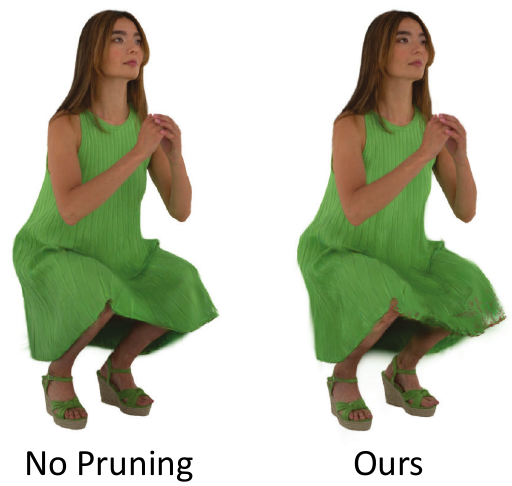}
  \end{center}
  \vspace{-6mm}
  \caption{Failure case of the pruning strategy. }
  \label{fig:limitation}
  \vspace{-1mm} 
\end{figure}

\section{Limitation}

For some challenging cases, \eg, an avatar wearing a loose garment, the pruned results may be worse than the unpruned ones, as shown in \cref{fig:limitation}. This may be because 20K blendshapes are still insufficient to represent such complex appearance. Keeping different numbers of blendshapes per dataset based on the variance magnitude might address this problem.

Additionally, since our application relies on WebGPU, whose support may be incomplete across different devices and browsers, we find that the application fails to run on some older devices, limiting broader application. Adopting a more mature graphics standard to implement may help address this issue.

\end{document}